\documentclass[twocolumn, switch]{article} 

\usepackage{preprint}
\usepackage{background}
\backgroundsetup{contents={}}

\usepackage{stfloats}
\usepackage{multirow}
\usepackage{pifont}
\usepackage{graphicx}
\usepackage{float}
\usepackage{afterpage}
\usepackage{caption}
\usepackage{hyperref}
\usepackage{url}

\usepackage{graphicx}
\usepackage{subcaption}

\usepackage{amsmath, amsthm, amssymb, amsfonts}

\usepackage[numbers,square]{natbib}
\usepackage[utf8]{inputenc}	
\usepackage[T1]{fontenc}	
\usepackage{xcolor}		

\usepackage{newfloat}
\DeclareFloatingEnvironment[name={Supplementary Figure}]{suppfigure}
\usepackage{sidecap}
\sidecaptionvpos{figure}{c}

\usepackage{titlesec}
\titlespacing\section{0pt}{12pt plus 3pt minus 3pt}{1pt plus 1pt minus 1pt}
\titlespacing\subsection{0pt}{10pt plus 3pt minus 3pt}{1pt plus 1pt minus 1pt}
\titlespacing\subsubsection{0pt}{8pt plus 3pt minus 3pt}{1pt plus 1pt minus 1pt}

\usepackage{tikz,xcolor,hyperref}

\definecolor{lime}{HTML}{A6CE39}
\DeclareRobustCommand{\orcidicon}{
	\begin{tikzpicture}
	\draw[lime, fill=lime] (0,0)
	circle [radius=0.16]
	node[white] {{\fontfamily{qag}\selectfont \tiny ID}};
	\draw[white, fill=white] (-0.0625,0.095)
	circle [radius=0.007];
	\end{tikzpicture}
	\hspace{-2mm}
}
\foreach \x in {A, ..., Z}{\expandafter\xdef\csname orcid\x\endcsname{\noexpand\href{https://orcid.org/\csname orcidauthor\x\endcsname}
			{\noexpand\orcidicon}}
}

\title{VAGNet: Vision-based accident anticipation with global features}

\usepackage{geometry}
\usepackage{eso-pic}
\usepackage{xcolor}

\AddToShipoutPictureBG*{%
  \AtPageLowerLeft{%
    \put(\LenToUnit{0.7in},\LenToUnit{0.5in}){
      \rule{2in}{0.5pt}}%
    \put(\LenToUnit{0.82in},\LenToUnit{0.33in}){
      \raggedright\textcolor{black}{\normalfont *correspondence: \hypersetup{pdfborder={0 0 0}}\href{mailto:charithc@cse.mrt.ac.lk}{\texttt{charithc@cse.mrt.ac.lk}}}}%
  }%
}

\usepackage{authblk}

\author[1]{Vipooshan Vipulananthan}
\author[1\thanks{\tt{charithc@cse.mrt.ac.lk}}]{Charith D. Chitraranjan}

\affil[1]{Department of Computer Science and Engineering, University of Moratuwa, Katubedda 10400, Sri Lanka (e-mail:
 \texttt{vipooshan.18@cse.mrt.ac.lk,
 charithc@cse.mrt.ac.lk}}

\usepackage{tikz}

\newcommand{\submittedtext}{%
\footnotesize\color{black}%
This is an open-access article published in IEEE Open Journal of Vehicular Technology (OJVT)). The final published version is available on IEEE Xplore at the following
DOI: 10.1109/OJVT.2026.3704953}

\newcommand{\topwatermark}{%
\begin{tikzpicture}[remember picture,overlay]
\node[anchor=north,yshift=-10pt] at (current page.north) {%
  \parbox{\dimexpr0.9\textwidth}{\submittedtext}%
};
\end{tikzpicture}%
}

\begin{document}

\twocolumn[ 
  \begin{@twocolumnfalse} 

\maketitle

\topwatermark

\begin{abstract}

Traffic accidents are a leading cause of fatalities and injuries across the globe. Therefore, the ability to anticipate hazardous situations in advance is essential. Automated accident anticipation enables timely intervention through driver alerts and collision avoidance maneuvers, forming a key component of advanced driver assistance systems. In autonomous driving, such predictive capabilities support proactive safety behaviors, such as initiating defensive driving and human takeover when required. Using dash-cam video as input offers a cost-effective solution, but it is challenging due to the complexity of real-world driving scenes. Accident anticipation systems need to operate in real-time. However, current methods involve extracting features from each detected object, which is computationally intensive. We propose VAGNet, a deep neural network that learns to predict accidents from dash-cam video using global features of traffic scenes without requiring explicit object-level features. The network consists of transformer and graph modules, and we use the vision foundation model VideoMAE-V2 for global feature extraction. Experiments on four benchmark datasets (DAD, DoTA, DADA, and Nexar) show that our method anticipates accidents with higher average precision while being computationally more efficient compared to existing methods. It also achieves higher mean time-to-accident scores in most cases, while being competitive in others.

\end{abstract}
\vspace{0.35cm}

  \end{@twocolumnfalse} 
] 



\section{Introduction}
Traffic accidents cost more than one million lives annually \cite{WHO_accidents, mannering2016unobserved}. Early and accurate anticipation by automated systems can mitigate accidents by warning the driver and, if necessary, initiating collision avoidance maneuvers. Therefore, it plays a critical role in advanced driver assistance systems (ADAS). In autonomous vehicles (AVs), it can proactively trigger a defensive driving mode in preparation for collision avoidance \cite{karim2022toward, zhang2026intelligent, mattas2022driver, li2024smpc, liu2025seeing} or, if applicable, for human takeover \cite{tekkesinoglu2025advancing}. AVs are primarily trained using reinforcement or imitation learning, and they cannot be exposed to all types of hazardous events that can occur in the real-world~\cite{chib2023recent}. This is commonly known as the "long-tail challenge" for AV safety~\cite{liu2024curse, zhang2023deep}. Therefore, as an additional layer of safety, it is important to have an accident anticipation system trained with a large number of diverse real-world accidents, such as those captured by dash-cams~\cite{moura2025nexar, yao2022dota, fang2019dada, yu2021scene}. 

Systems that use dash-cam video for accident anticipation are both cost-effective and easy to deploy. They can even be retrofitted to existing vehicles, which improves their safety~\cite{IRB_Retrofit, yoffie2014mobileye}. However, relying on vision is challenging due to the complex nature of driving scenes \cite{fang2023vision}. Over the last few years, many vision-based approaches utilizing dash-cam input have been proposed \cite{liu2025ccaf, zhongearly2025, thakur2024graph,karim2022dynamic, malawade2022spatiotemporal,bao2021drive, suzuki2018anticipating}. However, these methods involve extracting features from each detected object (usually VGG16~\cite{simonyan2014very} features from up to 19 objects), which is computationally demanding and can affect real-time performance. In this research, we avoid explicit object-level feature extraction and rely only on the global features of video frames. Furthermore, we use VideoMAE-V2~\cite{wang2023videomae} (base model), a pre-trained foundation model, to extract spatiotemporal global features, which provides better performance than other feature extractors such as I3D~\cite{carreira2017quo} and SlowFast~\cite{Feichtenhofer2018SlowFastNF} used in previous work. Our hypothesis is that the ability of foundation models to implicitly learn object-related features~\cite{simeoni2025dinov3} is sufficient to replace explicit object detection and feature extraction. 

Moreover, most existing methods are not evaluated on unbiased datasets with a sufficient number of ego-involved accidents. DAD~\cite{chan2017anticipating} and CCD~\cite{bao2020uncertainty} are the two dash-cam datasets primarily used in most studies. However, the former has very few ego-involved accidents, which is a critical limitation. The latter has a video quality bias between its positive and negative videos, making their discrimination a trivial task~\cite{mahmood2023new, karim2022dynamic}. In our evaluations, we use three benchmark datasets containing ego-involved accidents: DADA~\cite{fang2019dada}, DoTA~\cite{yao2022dota}, and Nexar~\cite{moura2025nexar}. We focus on ego-involved accidents, as they are the type that the ego-vehicle can attempt to mitigate directly, unlike those not involving the ego-vehicle. Nevertheless, we evaluated our proposed method on DAD as well for comparison with other work.

Our contributions can be summarized as follows.
\begin{itemize}
    \item We propose an efficient neural network architecture for accident anticipation that does not require explicit object detection or computationally expensive object-level feature extraction.
    \item We demonstrate the effectiveness of our approach with thorough experiments conducted on four benchmark datasets. 
\end{itemize}




\section{Literature Review}
In recent years, there has been growing research interest in vision-based accident anticipation, leading to the development of a wide range of new methods. The literature up to 2023 is reviewed in~\cite{fang2023vision, chitraranjan2025vision}. In this section, we review related work published from 2024 onward. 

Accident anticipation pipelines typically begin with object detection, followed by the extraction of object-level features from bounding boxes and global features from the full frame. Pretrained models are commonly used for object detection (e.g., Faster R-CNN~\cite{ren2015faster}) and feature extraction (e.g., VGG16~\cite{simonyan2014very}, I3D~\cite{carreira2017quo}). The resulting features are subsequently processed by neural networks with various architectural components to predict accident probabilities.

In the nested graph structure proposed by Thakur et al.~\cite{thakur2024graph}, the detected objects and their features are represented as nodes in a graph (object-level graph), where spatial relationships are modeled by edges between objects in the same frame, and temporal relationships are modeled by edges from objects in previous frames. Then, this object-level graph is pooled to produce a frame-level graph, where the nodes represent video frames. Furthermore, another frame-level graph is built using the global features of the frames. Unlike previous methods, they use I3D~\cite{carreira2017quo} to generate these global features so that the spatiotemporal dynamics of the video are captured. 

Finally, these two frame-level graphs are summarized with graph attention and fully connected layers to generate the probability of an accident. This approach was further improved in~\cite{vipulananthan2025stagnet} by using SlowFast networks~\cite{Feichtenhofer2018SlowFastNF} for global feature extraction and the addition of long short-term memory units (LSTMs) for direct temporal modeling. Another architecture with GNNs is proposed in~\cite{zhongearly2025}, where object and global features are combined using an attention-based fusion module. Subsequently, the combined features are used to construct a directed graph representing the temporal dependencies across frames. 

Several approaches have utilized transformers with multi-head attention. The AAT-DA~\cite{kumamoto2025aat} model employs a dual transformer architecture\textemdash a spatial and a temporal transformer\textemdash to  anticipate accidents. Furthermore, it integrates driver attention~\cite{zhao2023gated} when computing attention weights for detected objects, allowing the model to capture the influence of each object on surrounding objects. The methods proposed in~\cite{li2024cognitive, fang2022cognitive} fuse video data, textual descriptions, and driver attention maps using transformer-based architectures. However, they rely on human-annotated textual descriptions of accident videos during training. CCAF-Net~\cite{liu2025ccaf} integrates 3D depth information estimated from RGB images with object-level and global visual features. These multi-modal information streams are combined using collaborative attention, enabling the model to capture complex spatial relationships. However, computing depth information incurs an additional computing cost.

Although accident anticipation is expected to operate in real-time, the computational efficiency of most deep learning-based methods has not been evaluated~\cite{fang2023vision, chitraranjan2025vision}. A few recent methods have reported the number of model parameters, floating-point operations (FLOPs), or inference time\footnote{None of these studies reports all of these metrics.}~\cite{zhongearly2025, wang2023gsc, malawade2022spatiotemporal}. Note that most of the computational cost of these systems is due to the object detection and feature extraction backbones rather than the prediction model itself. The RARE framework~\cite{song2025real} addresses this challenge by using intermediate features from a pretrained object detector (YOLOv10~\cite{wang2024yolov10}), while avoiding more complex feature extractors. Other options include using lightweight versions of backbones such as YOLO-Tiny~\cite{falaschetti2024embedded, guan2022lightweight}. However, this comes with a trade-off in accuracy compared to more advanced architectures. 

Several methods have used hardware acceleration to speed up ADAS functions. Note that these systems are not designed to anticipate accidents although they are capable of object recognition, and multi-object, lane and drivable area detection and segmentation. For example, the work proposed in~\cite{tatar2023real} employs a ResNet-18~\cite{he2016deep} encoder and a Single Shot Detector architecture and implements it on the Kria KV260 MPSoC-FPGA platform. It uses hardware  optimization, where input activation, feature maps, and output metadata buffering are handled in on-chip memory. It also uses software optimization to promote data reuse and minimize the amount of external memory bandwidth required. Nguyen et. al~\cite{nguyen2024optimizing} utilize AI acceleration techniques available on the embedded platform, Jetson AGX Xavier, to achieve real-time performance. These include the TensorRT framework, multi-threaded programming, and post-training quantization.

However, the use of advanced spatio-temporal feature extractors while minimizing the dependence on additional features has not been explored yet. Furthermore, to the best of our knowledge, most of the advanced deep learning models for vision-based accident anticipation have not been adequately tested on edge devices.

\section{Methodology}

Extracting global contextual information together with object motion across video frames is essential for achieving a comprehensive understanding of dynamic scenes. Most existing methods~\cite{bao2020uncertainty, mahmood2023new, karim2022dynamic} rely on pretrained image-based networks, such as VGG16~\cite{simonyan2014very}, to obtain global features. However, these approaches are limited to spatial representations and do not model temporal dependencies across frames. Thakur et al.~\cite{thakur2024graph} employ the I3D network~\cite{carreira2017quo} to extract global frame-level features from short video clips, enabling joint spatial-temporal modeling. Nevertheless, such features remain inadequate for effectively capturing object interactions over time. To address these limitations, we adopt a pretrained VideoMAE-V2 network~\cite{wang2023videomae} to extract robust frame-level spatio-temporal representations.

\begin{figure*}[t!] 
    \centering
    \includegraphics[width=\textwidth, height=0.75\textheight, keepaspectratio]{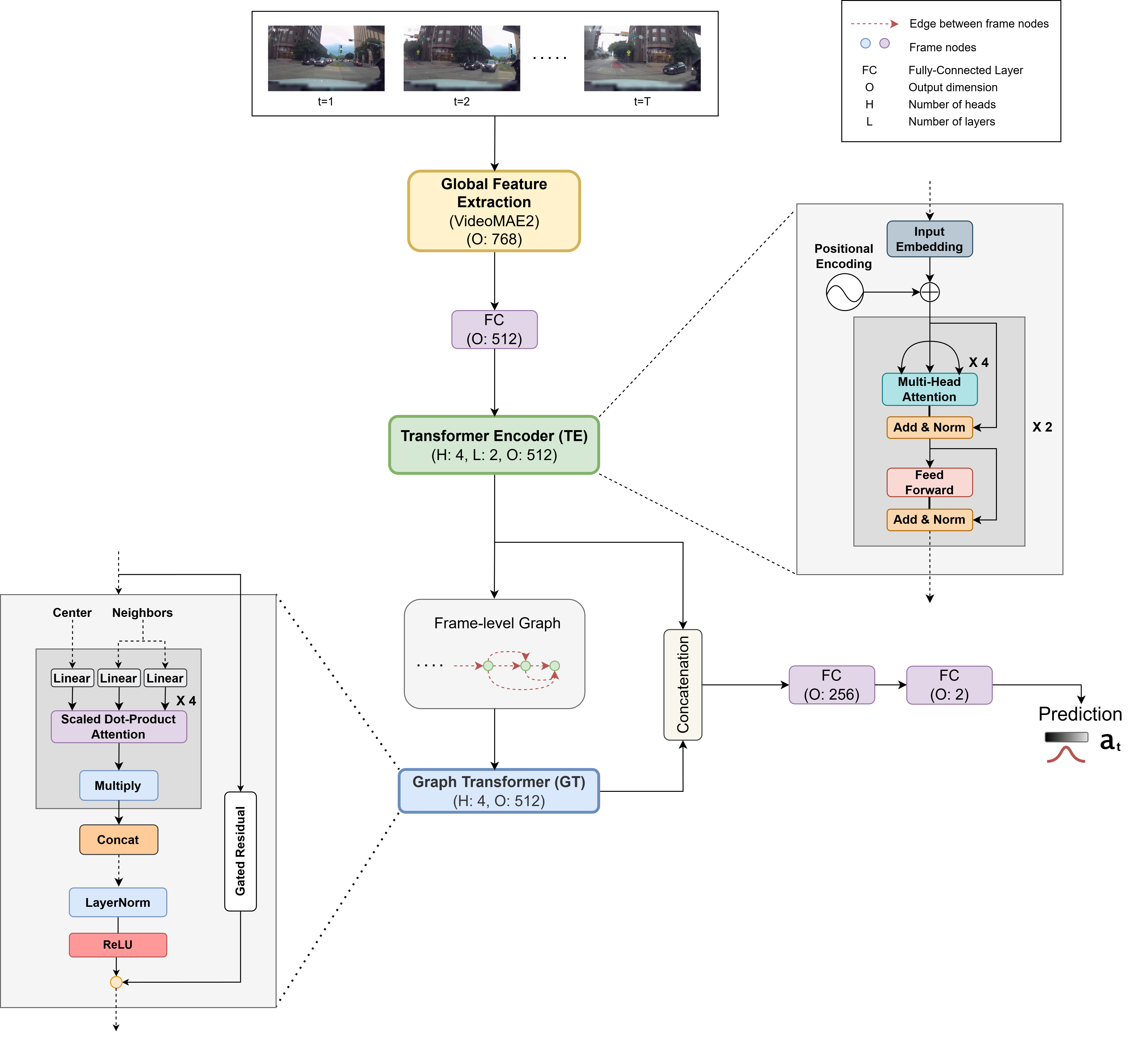}
    \caption{Architecture of the proposed VAGNet accident anticipation framework.}
    \label{fig1}
\end{figure*}

Figure~\ref{fig1} shows the overall architecture of our method. To obtain the global feature representation \( f_{\text{fr}}^t \) for a frame at time step \( t \), we input the current frame along with the preceding \(u\) frames, denoted as
\( X_{\text{seq}} = (X_{t-u}, \dots, X_t) \), into the VideoMAE-V2 network. This produces a high-dimensional spatio-temporal feature embedding, which is subsequently projected into a lower-dimensional space using a fully connected (FC) layer. To further strengthen the modeling of temporal dependencies, the reduced features are passed through a Transformer Encoder:

\setcounter{equation}{3}
\begin{equation}
\begin{aligned}
f_{\text{fr}} &= \mathrm{VideoMAE}(X_{\text{seq}}), \\
f^{\text{fc}}_{\text{fr}} &= \mathrm{FC}(f_{\text{fr}}), \\
f'_{\text{fr}} &= \mathrm{TE}(f^{\text{fc}}_{\text{fr}}),
\end{aligned}
\end{equation}

\noindent\textbf{Transformer Encoder (TE):}
The encoder follows the standard architecture of Vaswani \textit{et al.}~\cite{vaswani2017attention}, 
using \(N=2\) stacked encoder layers and \(h=4\) attention heads. The encoding process is,

\begin{equation}
H_{0} = f^{\text{fc}}_{\text{fr}}, \qquad
H_{i} = \mathrm{EncoderLayer}(H_{i-1}), \; i=1,2,...,N
\end{equation}
where the final encoded representation is \(f'_{\text{fr}}=H_{N}\).\\

\noindent\textbf{Encoder Layer:}
Each layer consists of multi-head self-attention (MHA) followed by a position-wise fully-connected layer (FC), both with residual connections and layer normalization:

\begin{equation}
\begin{aligned}
\tilde{H}_{i} &= \mathrm{LayerNorm}\!\left(H_{i-1} + \mathrm{MHA}(H_{i-1})\right), \\
H_{i} &= \mathrm{LayerNorm}\!\left(\tilde{H}_{i} + \mathrm{FC}(\tilde{H}_{i})\right).
\end{aligned}
\end{equation}





\noindent\textbf{Multi-Head Self-Attention:}
Given an input representation $H \in \mathbb{R}^{n \times d_{\text{model}}}$,
the multi-head self-attention (MHA) is defined as
\begin{equation}
\mathrm{MHA}(H) = \mathrm{Concat}(\text{head}_1,\ldots,\text{head}_h)\, W^{O},
\end{equation}
where $h$ denotes the number of attention heads and
$W^{O} \in \mathbb{R}^{d_{\text{model}}\times d_{\text{model}}}$ is a
learnable output projection matrix. Note that in this case, the sequence length $n=u+1$ (the number of input frames) and $d_{model}$ corresponds to the dimensionality of $f^{\text{fc}}_{\text{fr}}$.\\

Each attention head is computed using scaled dot-product attention:
\begin{equation}
\text{head}_j =
\mathrm{Softmax}\!\left(
\frac{Q_j K_j^{T}}{\sqrt{d_s}}
\right) V_j,
\quad j = 1,\ldots,h,
\end{equation}
where $d_s = d_{\text{model}} / h$ denotes the dimensionality
of each attention head (subspace dimension).

The query, key, and value matrices for the $j$-th head are obtained via
linear projections of $H$:
\begin{equation}
Q_j = H W_j^{Q}, \quad
K_j = H W_j^{K}, \quad
V_j = H W_j^{V},
\end{equation}
where $W_j^{Q}, W_j^{K}, W_j^{V} \in
\mathbb{R}^{d_{\text{model}} \times d_s}$ are learnable
projection matrices.

\noindent\textbf{Fully-Connected Layer (FC):}

\begin{equation}
\mathrm{FC}(x) = \sigma(xW_{1}+b_{1})W_{2}+b_{2},
\end{equation}

where \(\sigma(\cdot)\) denotes a non-linear activation function (ReLU).
The resulting encoded representation is subsequently integrated into a Graph Transformer module, as described below.

To explicitly model temporal relationships between frames, we construct a \textit{frame graph} using the global frame-level features obtained in the previous stage. Each node in the graph corresponds to a video frame at time \(t\), and directed edges connect each frame to its preceding \(v\) frames, ensuring a causal structure that prevents information leakage from future frames. The adjacency matrix \(A_{\text{fr}}\) is defined as:

\begin{equation}
A_{\text{fr}}(i, j) =
\begin{cases}
1, & \text{if } i - j \leq v, \\
0, & \text{otherwise},
\end{cases}
\end{equation}

where \(i\) denotes the index of the current frame node.

A Graph Transformer layer~\cite{shi2021maskedlabelpredictionunified} is then applied to process the global frame-level features \(f'_{\text{fr}}\):

\begin{equation}
\begin{aligned}
f''_{\text{fr}} &= \text{GT}(f'_{\text{fr}}, A_{\text{fr}}), \\
f'' &= [f'_{\text{fr}}, f''_{\text{fr}}],
\end{aligned}
\end{equation}

where the attention mechanism enables the model to identify temporally significant frames that contribute more effectively to accident prediction. The concatenated representation \(f''\) is subsequently passed through fully connected layers to estimate frame-level accident probabilities \((p_1, p_2, \dots, p_N)\).

The entire framework is trained end-to-end, except for the VideoMAE-V2 global feature extraction network, which remains frozen during training. Optimization is performed using the standard cross-entropy loss:

\begin{equation}
\mathcal{L}(p, y) = - \sum_{m=1}^{M} y_m \log \left( \frac{e^{p_m}}{\sum_j e^{p_j}} \right),
\end{equation}

where \(M\) denotes the number of classes (two in our case), \(p_m\) represents the predicted logits, and \(y_m\) is the ground-truth label indicator.

\section{Experimental Evaluation}

\subsection{Experimental Settings}

\subsubsection{Datasets}

We evaluate our approach on four benchmark datasets: DAD \cite{chan2017anticipating}, DADA \cite{fang2019dada}, DoTA \cite{yao2022dota}, and Nexar \cite{moura2025nexar}. For DAD, we adopt the same training and testing splits as prior studies \cite{bao2020uncertainty, thakur2024graph, mahmood2023new}. However, the majority of accidents in DAD do not involve the ego vehicle. Nevertheless, ego-involved accidents are the most critical for Advanced Driver Assistance Systems (ADAS) since these are the events that such systems can actively prevent or mitigate. Therefore, we focus on ego-involved accidents in the DADA, DoTA, and Nexar datasets.

All accident instances in the Nexar dataset~\cite{moura2025nexar} are ego-involved. Therefore, we utilize the entire dataset for our evaluation with its original training/testing splits. To ensure consistency with our implementation and the other benchmarks, training videos are trimmed to 5-second clips so that, in positive samples, the accident occurs at the end of the clip. In addition, all videos are down-sampled to 10 frames per second.

\begin{figure}[!t]
    \centering
    \includegraphics[width=\columnwidth]{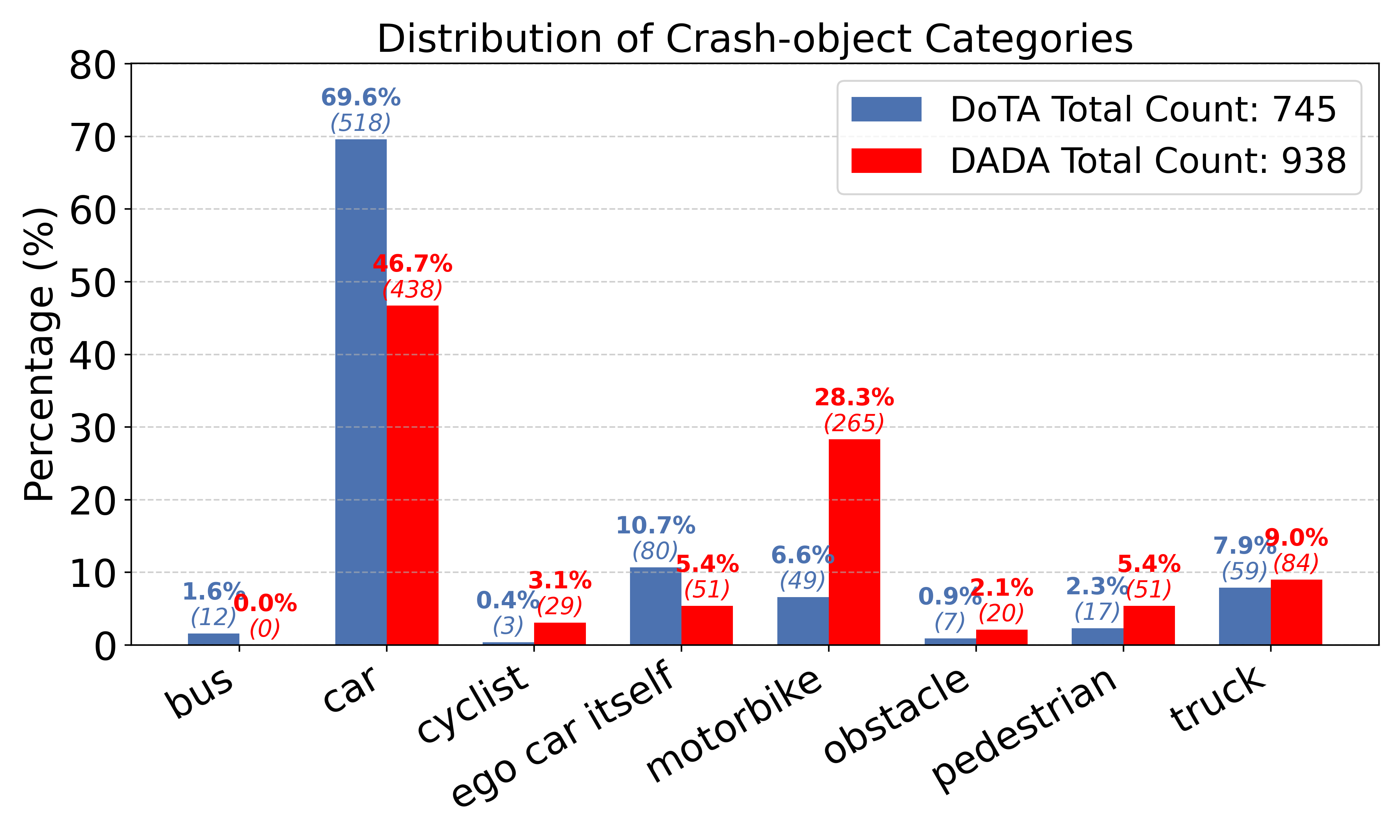}
    \caption{Distribution of crash-object categories in the DoTA and DADA datasets. Crash-object category annotations are not available for the Nexar dataset.}
    \label{fig:dada_co_dist}
\end{figure}

For DADA and DoTA, we select videos annotated as ego-involved accidents and extract 5-second clips such that the accident takes place within the final 2 seconds, with the exact onset randomly shifted. These clips constitute the positive samples. Unlike DAD and Nexar, DADA and DoTA contain only accident videos. To construct unbiased negative samples, we identify videos with sufficient normal driving footage prior to the accident and extract 5-second segments exclusively from these non-accident portions. This sampling strategy is similar to the protocol adopted by Mahmood et al. \cite{mahmood2023new}. Dataset statistics are summarized in Table \ref{tab:dataset_stats}. While all available ego-involved accident videos from DADA (938 samples) are included, only 745 videos are selected from DoTA. This is because only 745 suitable negative clips could be extracted from DoTA given the available video durations and accident onsets. To maintain class balance, an equal number of positive samples is used, consistent with prior work \cite{mahmood2023new}. Beyond simplifying optimization, maintaining a balanced dataset avoids artificially inflated average precision scores caused by class imbalance \cite{saito2015precision}. Since standard training and testing splits are not available for the DADA and DoTA datasets, we employed five-fold cross-validation for effective utilization of the data. To prevent any information leakage, video clips originating from the same source video were grouped within the same fold during cross-validation. Figure~\ref{fig:dada_co_dist} shows the distribution of the crash-objects in DoTA and DADA accident videos we have used.

For each accident dataset, normal driving videos are sourced from the same dataset rather than from external collections. This design choice prevents discrepancies in visual quality between positive and negative samples that models could exploit to achieve deceptively high performance without learning meaningful accident-related patterns \cite{mahmood2023new, karim2022dynamic, chitraranjan2025vision}. For the same reason, we exclude the widely used CCD dataset \cite{bao2020uncertainty}. In CCD, accident videos are predominantly low-quality clips collected from YouTube, whereas normal driving videos are high-quality samples from BDD100k \cite{yu2020bdd100k}. As a result, many prior methods report average precision values exceeding 99\%, likely due to this quality bias rather than genuine accident anticipation capability, making CCD unsuitable for fair comparative evaluation.

\begin{table}[htbp]
\caption{Dataset Statistics.}
\centering
\begin{tabular}{|l|c|c|c|}
\hline
\textbf{Dataset} & \textbf{Positive} & \textbf{Negative} & \textbf{FPS} \\
\hline
DAD~\cite{chan2017anticipating} & 620 & 1130 &  20 \\
DoTA~\cite{yao2022dota} & 745 & 745 & 10 \\
DADA~\cite{fang2019dada} & 938 & 1361 & 10 \\
Nexar~\cite{moura2025nexar} & 1422 & 1422 & 10 \\
\hline
\end{tabular}
\label{tab:dataset_stats}
\end{table}

\subsubsection{Networks and Baselines}

For global scene modeling, we employ the VideoMAE-V2\cite{wang2023videomae} architecture, which produces a frame-level global feature representation of dimension $h = 768$. The network takes the current frame along with the preceding $u=15$ frames as input. For frame-level graph construction, the number of temporal neighbors is set to $v = 20$.

VAGNet was implemented using PyTorch 2.0.0 and trained for 20 epochs using the Adam optimizer with a learning rate of $1\times10^{-4}$ and a batch size of 1. This same training configuration was applied across all four datasets.

Furthermore, we executed the following comparison methods on the DoTA, DADA, and Nexar datasets: UString~\cite{bao2020uncertainty}, UString(Geometrics), Graph(Graph)~\cite{thakur2024graph}, AAT-DA~\cite{kumamoto2025aat}, and StagNet~\cite{vipulananthan2025stagnet}.

For UString~\cite{bao2020uncertainty} and Graph(Graph)~\cite{thakur2024graph}, we used the original implementations made publicly available by their respective authors. We implemented UString (Geometric) as an approximation of GSTL~\cite{mahmood2023new} by substituting the VGG16-based object-level features used in UString~\cite{bao2020uncertainty} with geometric features computed from detected bounding boxes.

As the source code for AAT-DA is not publicly available, we re-implemented the method following the model configuration reported in the original paper. Specifically, we adopted the feature extraction pipeline commonly used in prior work~\cite{bao2020uncertainty, karim2022dynamic}, considering the top 19 detected objects and extracting 4096-dimensional object-level and global features using VGG16. The hidden dimension and the number of attention heads were set to 1024 and 8, respectively, for all transformer modules. Dropout was applied to the spatial transformer, object self-attention layer, temporal transformer, and fully connected layer, with rates of 0.3, 0.3, 0.1, and 0.5, respectively. Note that for the DAD dataset, our implementation achieved very similar results to those reported in the original work (63\% vs. 64\% AP), suggesting a reasonable re-implementation.


All experiments were conducted on a system equipped with an NVIDIA Tesla T4 GPU featuring 15 GiB GDDR6 VRAM and 320 GB/s memory bandwidth. The runtime environment also included an Intel Xeon CPU operating at 2.00 GHz with approximately 12 GiB of system RAM. This provides a resource-constrained setup, comparable to edge AI platforms such as Nvidia Jetson AGX Orin or Jetson Thor. For the evaluation of deployment-oriented inference, models were exported to the ONNX format and optimized using NVIDIA TensorRT with FP16 precision. This was done through the ONNX Runtime TensorRT Execution Provider to emulate inference in edge-device environments.

\subsubsection{Evaluation Metrics}
\noindent

Our objective is to predict traffic accidents as accurately and as early as possible. Earlier and more reliable predictions provide greater potential for accident prevention. At each video frame, the model outputs a risk score that reflects the probability of an accident occurring in the future. If this score exceeds a predefined decision threshold, the frame is classified as indicating an impending accident.

To evaluate prediction accuracy, we adopt the Average Precision (AP) metric, which is derived from precision and recall. Precision ($P$) and recall ($R$) are defined as:
\begin{equation}
P = \frac{TP}{TP + FP},
\end{equation}
\begin{equation}
R = \frac{TP}{TP + FN},
\end{equation}
where $TP$, $FP$, and $FN$ denote true positives, false positives, and false negatives, respectively. The Average Precision is computed as the integral of precision with respect to recall as the decision threshold is varied:
\begin{equation}
AP = \int P_R \, dR,
\end{equation}
where $P_R$ represents the precision corresponding to a given recall value $R$.

In addition to accuracy, we assess the earliness of accident prediction using the mean Time-to-Accident (mTTA) metric. When the predicted risk score is greater than or equal to the threshold, the model assumes that an accident will occur in the future. Time-to-Accident (TTA) is defined as the time interval between the earliest frame at which the predicted score exceeds the threshold and the actual accident onset:
\begin{equation}
\mathrm{TTA} = \max \{\tau - t \mid a_t \geq \hat{a}, \, 1 \leq t \leq \tau\},
\end{equation}
where $\tau$ denotes the time at which the accident begins, $a_t$ is the predicted risk score at frame $t$, and $\hat{a}$ is the decision threshold.

Since TTA varies with the decision threshold ($\hat{a}$), multiple TTA measurements can be obtained by adjusting the threshold. The mean Time-to-Accident is therefore computed as:
\begin{equation}
\mathrm{mTTA} = \mathbb{E}[\mathrm{TTA}].
\end{equation}

Compared to TTA measured at a fixed threshold, mTTA provides a more accurate assessment of early prediction performance. A larger mTTA value indicates that the model is capable of predicting accidents earlier, reflecting better early-warning capability.

\begin{figure*}[!t]
\centering

\includegraphics[width=0.95\textwidth]{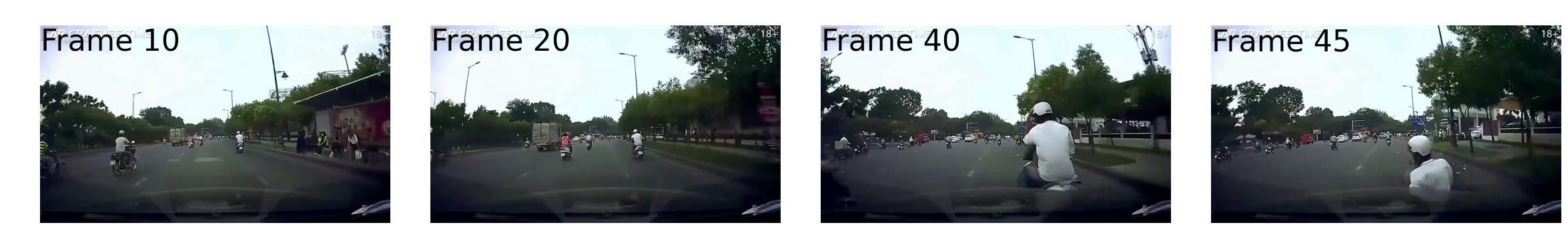}
\vspace{-4pt}

\includegraphics[width=0.95\textwidth]{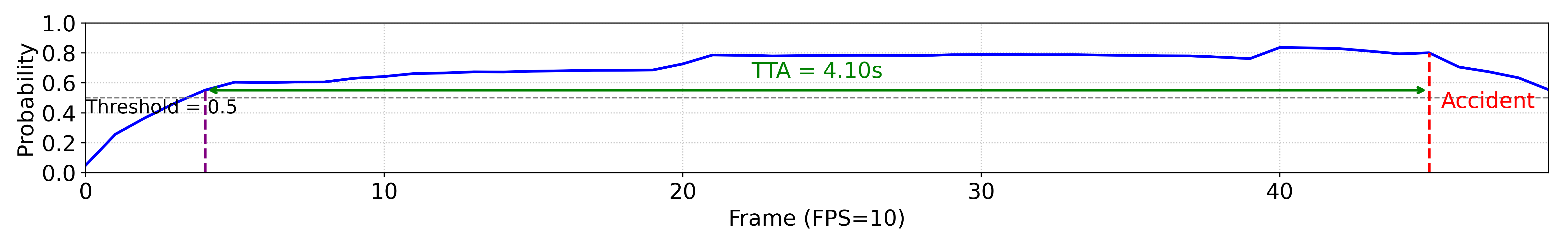}
\vspace{-6pt}

{\footnotesize (a) Accident (true positive) - Daytime, cloudy weather}
\vspace{6pt}

\includegraphics[width=0.95\textwidth]{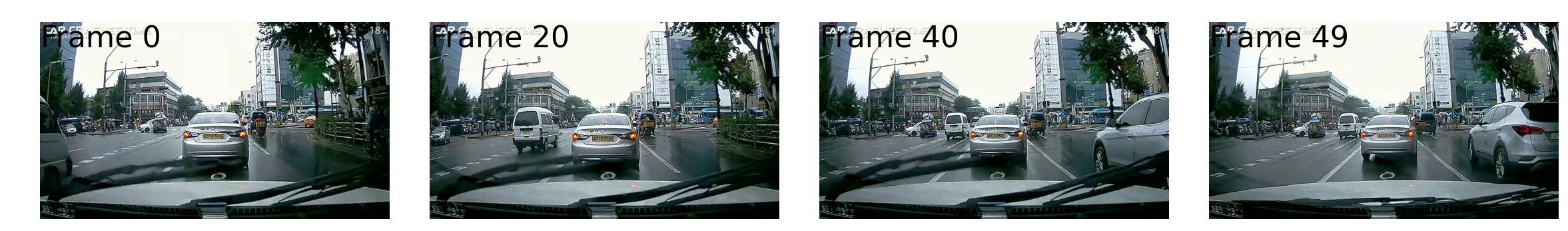}
\vspace{-4pt}

\includegraphics[width=0.95\textwidth]{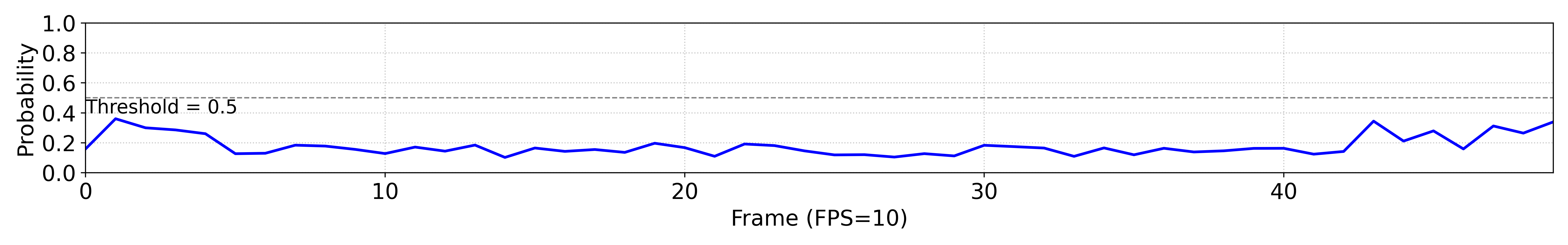}
\vspace{-6pt}

{\footnotesize (b) Non-accident (true negative) - Daytime, rainy weather}
\vspace{6pt}

\includegraphics[width=0.95\textwidth]{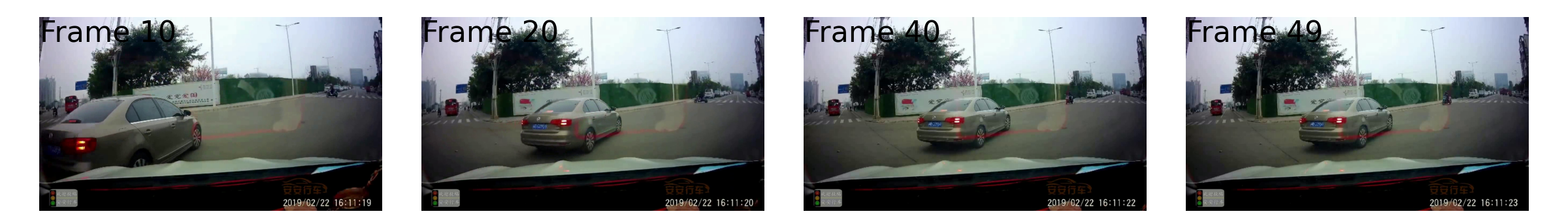}
\vspace{-4pt}

\includegraphics[width=0.95\textwidth]{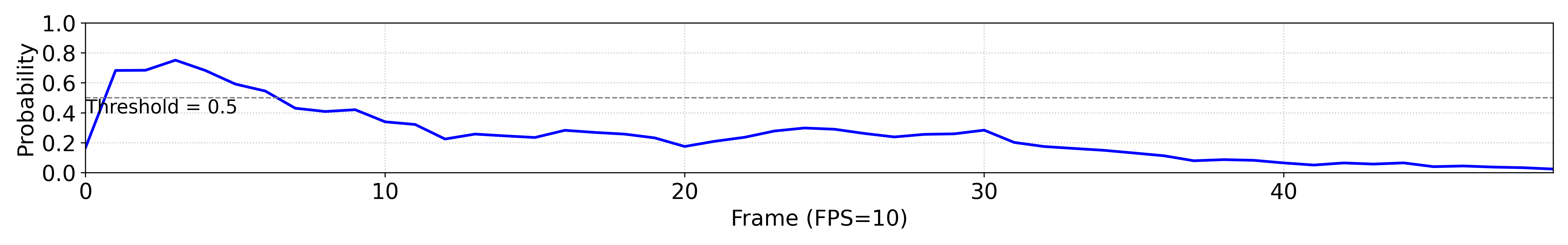}
\vspace{-6pt}

{\footnotesize (c) Non-accident (false positive) - Daytime, clear weather}
\vspace{6pt}

\includegraphics[width=0.95\textwidth]{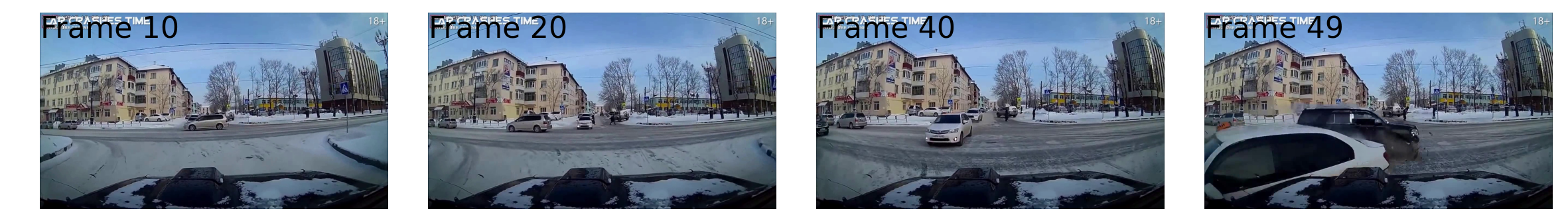}
\vspace{-4pt}

\includegraphics[width=0.95\textwidth]{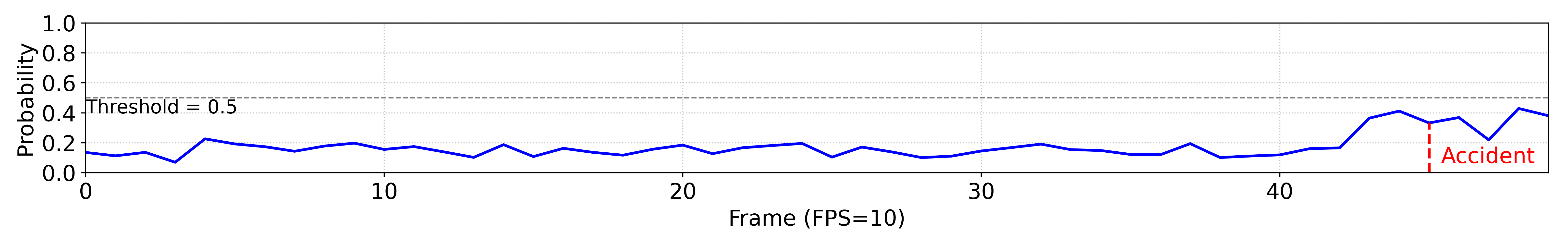}
\vspace{-6pt}

{\footnotesize (d) Accident (false negative) - Daytime, snowy weather}

\caption{Samples showing the variation of predicted accident probability across time/frames for (a) True positive (b) True negative (c) False positive and (d) False negative cases.}
\label{fig:frame_prob}
\end{figure*}

\begin{figure*}[!tb]
\centering

\vspace{-2mm}

\begin{minipage}{0.88\textwidth}
    \centering
    \includegraphics[width=\textwidth]{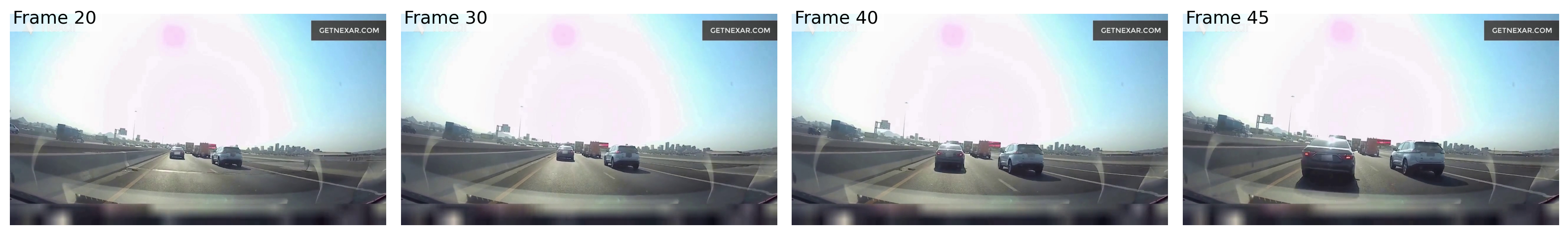}
\end{minipage}

\vspace{1pt}

\begin{minipage}{0.88\textwidth}
    \centering
    \includegraphics[width=\textwidth]{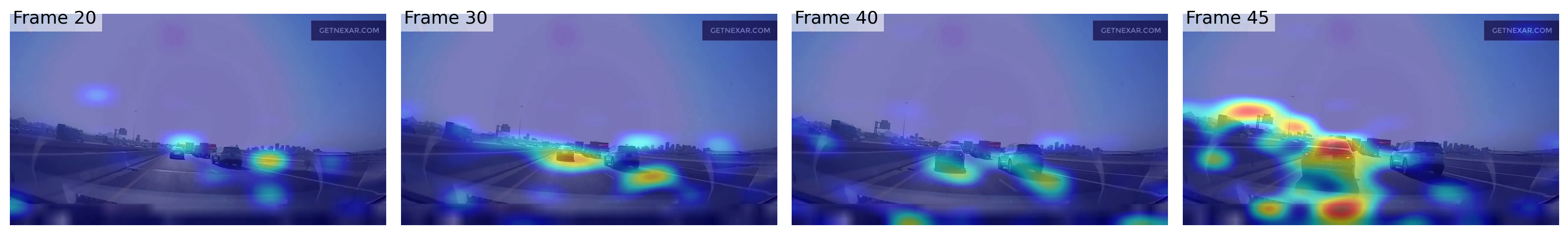}
    \vspace{-4pt}
    {\footnotesize (a) Daytime, clear weather}
\end{minipage}

\vspace{8pt}

\begin{minipage}{0.88\textwidth}
    \centering
    \includegraphics[width=\textwidth]{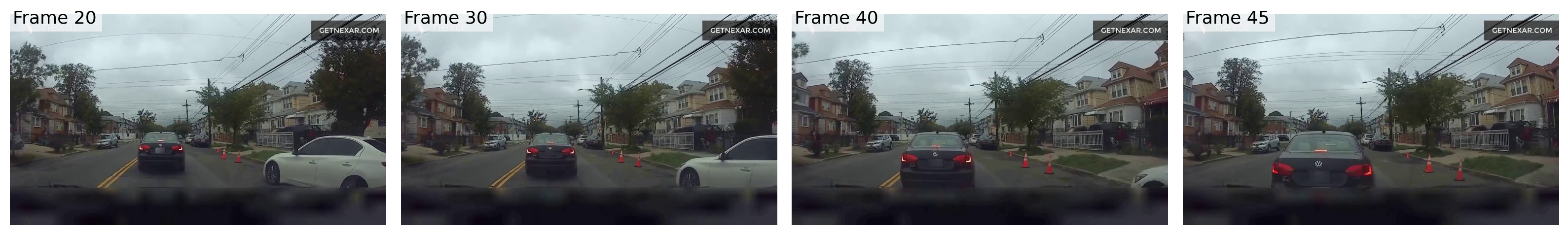}
\end{minipage}

\vspace{1pt}

\begin{minipage}{0.88\textwidth}
    \centering
    \includegraphics[width=\textwidth]{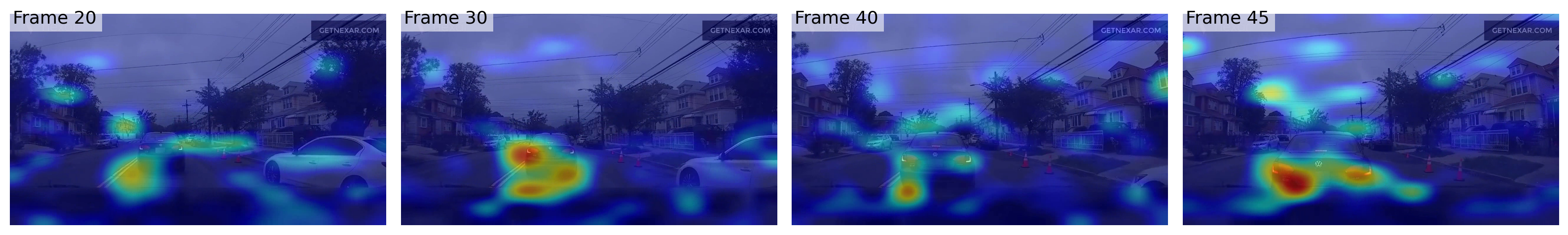}
    \vspace{-4pt}
    {\footnotesize (b) Daytime, cloudy weather}
\end{minipage}

\vspace{8pt}

\begin{minipage}{0.88\textwidth}
    \centering
    \includegraphics[width=\textwidth]{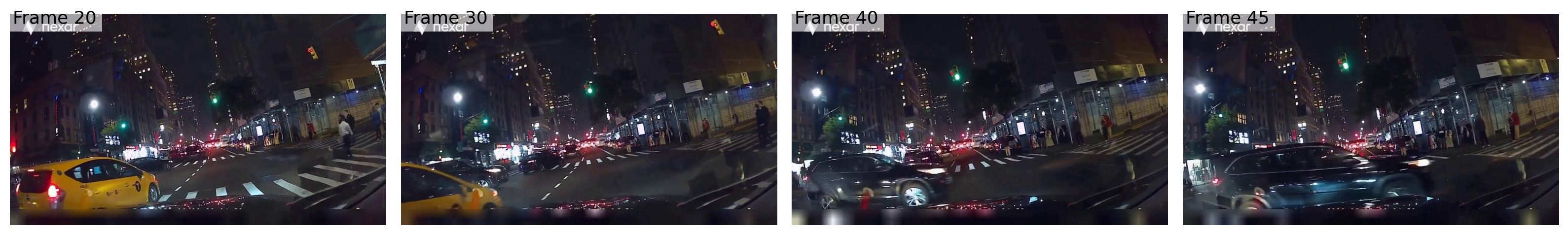}
\end{minipage}

\vspace{1pt}

\begin{minipage}{0.88\textwidth}
    \centering
    \includegraphics[width=\textwidth]{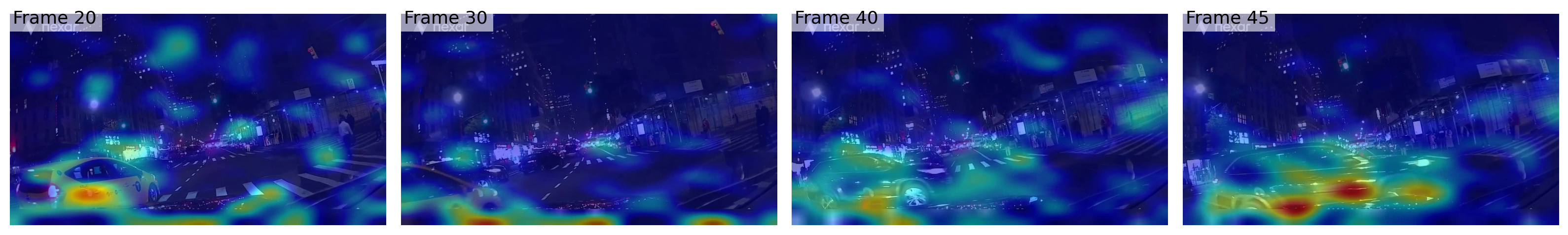}
    \vspace{-4pt}
    {\footnotesize (c) Nighttime, clear weather}
\end{minipage}

\vspace{8pt}

\begin{minipage}{0.88\textwidth}
    \centering
    \includegraphics[width=\textwidth]{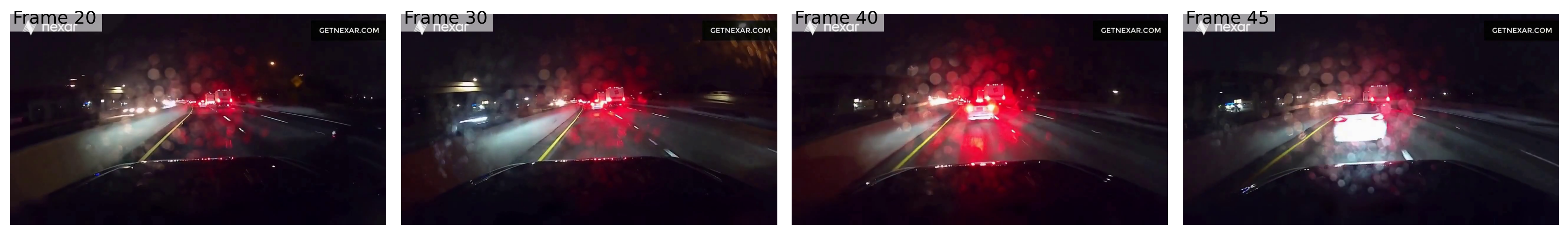}
\end{minipage}

\vspace{1pt}

\begin{minipage}{0.88\textwidth}
    \centering
    \includegraphics[width=\textwidth]{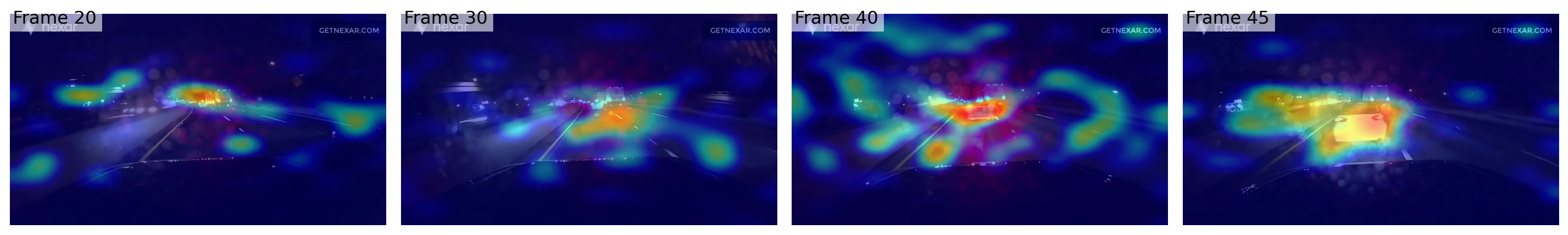}
    \vspace{-4pt}
    {\footnotesize (d) Nighttime, rainy weather}
\end{minipage}

\vspace{2pt}

\caption{\textbf{Samples of visual explanations for accidents across different weather/lighting conditions.} For each condition, the top row shows the original video frames in temporal order, and the bottom row shows the corresponding Grad-CAM~\cite{selvaraju2017grad} maps highlighting areas in the image that contribute to accident anticipation. Higher intensity indicates regions that have a stronger influence on the model’s anticipation of a potential accident.}
\label{fig:grad_cam}

\vspace{-2mm}
\end{figure*}

\subsection{Results and Discussion}

\subsubsection{Comparison with State-of-the-art Methods}

As shown in Table~\ref{tab:results_comparison}, the proposed VAGNet model achieves higher AP and mTTA than existing methods across all four datasets, except for the marginally higher mTTA attained by AAT-DA and Graph(Graph) on DoTA and DAD, respectively. All results corresponding to prior work on DAD were extracted from their respective publications. Results on the other three datasets were obtained by executing the models as described earlier. Unlike our method, all prior work that we have compared uses object detection and object-level features, while some even use additional information such as depth (CCAF-Net~\cite{liu2025ccaf}) and driver attention (AAT-DA~\cite{kumamoto2025aat}), all of which contribute to the computational cost. Our results indicate that the effective incorporation of global features is sufficient to outperform these methods while being more efficient. A likely reason for this is the capability of vision foundation models to implicitly learn object boundaries, as demonstrated by their state-of-the-art performance on tasks such as object detection and segmentation~\cite{simeoni2025dinov3}.

Figure~\ref{fig:frame_prob} shows samples of how the models' predicted accident probabilities vary over time. The true positive case gives an example of how the model identifies an accident early. The false positive/negative examples demonstrate the challenging nature of traffic scenes, where an accident and a near-pass appear very similar from the dash-cam view. 

We also analyzed the frames preceding accidents using Grad-Cam to visualize which areas of the input frame contribute more to the models predictions. As demonstrated by the sample images in Figure~\ref{fig:grad_cam}, the model tends to focus on the nearby objects, as expected.

We conducted additional experiments to analyze the contributions of the individual components of VAGNet to its predictive performance, including the effects of different global feature extraction backbones, the transformer encoder, and the GNN modules.

\noindent

\begin{table}[htbp]
\caption{Performance comparison of different methods.}
\centering
\scriptsize
\begin{tabular}{|c|c|c|c|}
\hline
\textbf{Dataset} & \textbf{Method} & \textbf{AP (\%)} & \textbf{mTTA (s)} \\
\hline
\multirow{4}{*}{DAD~\cite{chan2017anticipating}} 
                & DSA (2017)~\cite{chan2017anticipating}       & 48.1 & 1.34 \\
                & adaLEA (2018)~\cite{suzuki2018anticipating}   & 52.3 & 3.43 \\
                & UString (2020)~\cite{bao2020uncertainty}  & 53.7 & 3.53 \\
                & DSTA (2022)~\cite{karim2022dynamic}  & 56.1 & 3.66 \\
                & EGSTL (2023)~\cite{mahmood2023new}       & 56.3 & 3.23 \\
                & GSC (2024)~\cite{wang2023gsc}       & 60.4 & 3.53 \\
                & DAA-GNN (2024)~\cite{song2024dynamic}       & 75.2 & 1.47 \\
                & Graph(Graph) (2024)~\cite{thakur2024graph}       & 63.6 & \textbf{4.45} \\
                & RARE (2025)~\cite{song2025real}       & 62.2 & \textbf{2.51} \\
                & AAT-DA (2025)~\cite{kumamoto2025aat}       & 64.0 & 2.87 \\
                & STAGNet (2025)~\cite{vipulananthan2025stagnet}        & 64.36 & 4.32 \\
                & CCAF-Net (2025)~\cite{liu2025ccaf} & 71.8 & 4.15\\
                & Zhong et. al (2025)~\cite{zhongearly2025} & 72.7 & 3.98\\
                & VAGNet (Ours)        & \textbf{77.98} & 4.41 \\
\hline
\multirow{4}{*}{DoTA ~\cite{yao2022dota}} & UString (Geometrics) ~\cite{mahmood2023new} & 63.2 & 2.14 \\
                    & UString (2020)~\cite{bao2020uncertainty}                & 75.4 & 2.26 \\
                    & Graph(Graph) (2024)~\cite{thakur2024graph}               & 83.5 & 3.14 \\
                    & AAT-DA (2025)~\cite{kumamoto2025aat}                     & 72.4 & \textbf{3.28} \\
                    & STAGNet (2025)~\cite{vipulananthan2025stagnet}           &  92.73 & 3.23 \\
                    & VAGNet(Ours)                 &                \textbf{96.66} & 3.22 \\
\hline
\multirow{4}{*}{DADA ~\cite{fang2019dada}} & UString (Geometrics) ~\cite{mahmood2023new} & 56.8 & 1.25 \\
                      & UString  (2020)~\cite{bao2020uncertainty}            & 63.9 & 1.03 \\
                      & Graph(Graph) (2024)~\cite{thakur2024graph}                & 84.0 & 2.92 \\
                    & AAT-DA (2025)~\cite{kumamoto2025aat}                         & 69.4 & 1.36 \\
                    & STAGNet (2025)~\cite{vipulananthan2025stagnet}               & 95.39 & 3.08 \\
                    & VAGNet (Ours)                 & \textbf{97.82} & \textbf{3.09} \\
\hline
\multirow{4}{*}{Nexar ~\cite{moura2025nexar}} & 
                    UString (Geometrics) ~\cite{mahmood2023new} & 71.0 & 3.33 \\
                      & UString  (2020)~\cite{bao2020uncertainty}            & 73.3 & 2.66 \\
                      & Graph(Graph) (2024)~\cite{thakur2024graph}                & 60.4 & 4.61 \\
                    & AAT-DA (2025)~\cite{kumamoto2025aat}                         & 75.1 & 2.50 \\
                    & STAGNet (2025)~\cite{vipulananthan2025stagnet}               & 76.5 & 3.72 \\
                    & VAGNet (Ours)                 & \textbf{81.25} & \textbf{4.69} \\
\hline
\end{tabular}
\label{tab:results_comparison}
\end{table}

\subsubsection{Impact of Global Features}

Experimental results presented in Table \ref{tab:glob_ablation} demonstrate the impact of different global feature extraction backbones, namely VGG16, I3D, SlowFast, and VideoMAE-V2, across the four benchmark datasets. The results show a consistent improvement in Average Precision (AP) as the global feature representation evolves from the 2D CNN-based VGG16 to spatiotemporal models such as I3D and SlowFast, highlighting the importance of spatiotemporal modeling for accident anticipation. Among all evaluated backbones, VideoMAE-V2 attains the highest AP across all datasets, with substantial improvements on DAD and Nexar, highlighting its effectiveness in modeling  temporal and spatial dependencies in complex scene dynamics. VideoMAE-V2’s better performance can be largely attributed to its pretraining as a foundation model on diverse large-scale video datasets. Although SlowFast yields slightly higher mTTA values on DoTA and DADA, VideoMAE-V2 maintains competitive mTTA while considerably improving detection accuracy. Note that a few models using spatial (VGG16-based) global features have reached an AP close to that of those using more advanced spatiotemporal (SlowFast-based) global features. Specifically, CCAF-Net and AAT-DA have achieved AP scores that are very similar to those of STAGNet on the DAD and Nexar datasets, respectively. However, CCAF-Net performs monocular depth estimation, while AAT-DA computes driver attention, both of which require additional modules at an extra computational cost. 

Overall, these results demonstrate that effective video-level global representations play a critical role in enhancing both early and accurate accident anticipation.

\begin{table}[htbp]
\scriptsize
\setlength{\tabcolsep}{4pt}
\caption{Performance of VAGNet under different global feature extraction backbones.}
\centering
\begin{tabular}{|c|c|c|c|}
\hline
\textbf{Dataset} & \shortstack{\textbf{Global}\\ \textbf{Features}} & \textbf{AP (\%)} & \textbf{mTTA (s)} \\
\hline

\multirow{3}{*}{DAD}
 & VGG16        & 54.83 & 4.15 \\
 & I3D        & 59.27 & 4.13 \\
 & SlowFast  & 62.90 & 3.74 \\
 & VideoMAE-V2 & \textbf{77.98} & \textbf{4.41} \\
\hline

\multirow{3}{*}{DoTA}
 & VGG16        & 88.74 & 3.25 \\
 & I3D        & 91.71 & 3.20 \\
 & SlowFast  & 95.06 & \textbf{3.26} \\
 & VideoMAE-V2 & \textbf{96.66} & 3.22 \\
\hline

\multirow{3}{*}{DADA}
 & VGG16        & 83.61 & 3.12 \\
 & I3D        & 90.57 & 3.09 \\
 & SlowFast  & 95.83 & \textbf{3.18} \\
 & VideoMAE-V2 & \textbf{97.82} & 3.09 \\
\hline

\multirow{3}{*}{Nexar}
 & VGG16        & 62.89 & 4.48 \\
 & I3D        & 65.3 & 4.43 \\
 & SlowFast  & 73.89 & 4.67 \\
 & VideoMAE-V2 & \textbf{81.25} & \textbf{4.69} \\
\hline

\end{tabular}
\label{tab:glob_ablation}
\end{table}

\subsubsection{Ablation Study}

Table \ref{tab:ablation_combined} presents an ablation study investigating the individual contributions of the Transformer Encoder (TE) and the Graph Transformer (GT) across the DAD, DoTA, DADA, and Nexar datasets. Removing both components so that the model consists only of VideoMAE-V2 feature extraction followed by fully connected layers significantly degrades performance, resulting in lower AP and mTTA values across all datasets. This highlights the necessity of modeling temporal and relational information. Incorporating either the TE or the GT substantially improves AP compared to the baseline. The full model, incorporating both TE and GT modules, further improves performance, but to a lesser extent. Nevertheless, it consistently achieves the highest AP and comparable mTTA on all datasets. This suggests that TE and GT modules mostly capture similar aspects of accident dynamics while also capturing some complementary aspects. 

\begin{table}[htbp]
\scriptsize
\setlength{\tabcolsep}{4pt}
\caption{Results of the ablation study. Each model variant removes one or more components (Transformer Encoder or Graph) to analyze their individual impact.}
\centering
\begin{tabular}{|c|c|c|c|c|c|}
\hline
\textbf{Data} & \textbf{TE} & \textbf{GT} & \textbf{AP (\%)} & \textbf{mTTA (s)} \\
\hline
\multirow{7}{*}{DAD ~\cite{chan2017anticipating}} 
 & \ding{55} & \ding{55} & 64.15 & 2.54 \\
 & \ding{51} & \ding{55} & 76.58 & 4.35 \\
 & \ding{55} & \ding{51} & 71.37 & 4.12 \\
 & \ding{51} & \ding{51} & \textbf{77.98} & \textbf{4.41} \\
\hline
\multirow{7}{*}{DoTA ~\cite{yao2022dota}} 
 & \ding{55} & \ding{55} & 87.50 & 2.45 \\
 & \ding{51} & \ding{55} & 95.87 & \textbf{3.31} \\
 & \ding{55} & \ding{51} & 95.00 & 3.12 \\
 & \ding{51} & \ding{51} & \textbf{96.66} & 3.22 \\
\hline
\multirow{7}{*}{DADA ~\cite{fang2019dada}} 
 & \ding{55} & \ding{55} & 75.43 & 1.83 \\
 & \ding{51} & \ding{55} & 97.07 & \textbf{3.16} \\
 & \ding{55} & \ding{51} & 97.10 & 3.10 \\
 & \ding{51} & \ding{51} & \textbf{97.82} & 3.09 \\
\hline
\multirow{7}{*}{Nexar ~\cite{moura2025nexar}} 
 & \ding{55} & \ding{55} & 76.34 & 3.11 \\
 & \ding{51} & \ding{55} & 78.02 & \textbf{4.73} \\
 & \ding{55} & \ding{51} & 80.15 & 4.13 \\
 & \ding{51} & \ding{51} & \textbf{81.25} & 4.69 \\
\hline
\end{tabular}
\label{tab:ablation_combined}
\end{table}

\subsubsection{Analysis of Computational Efficiency}

Since accident anticipation needs to operate in real-time, we analyzed the computational cost of inference. In Tables~\ref{tab:execution_flops} and~\ref{tab:execution_fps}, we report the number of floating point operations (FLOPs) and time, respectively, needed to perform a forward pass per video frame during inference. Unlike most prior work, which reports only model inference costs, we also include the computational costs of the feature extraction backbones. As evident from our analysis, feature extraction from detected objects requires the most FLOPs, both in the worst case (19 objects/frame) and the average case (6.12 objects/frame), computed over the DoTA, DADA and Nexar datasets. It is also accounts for a large proportion of the inference time. The proposed VAGNet model avoids this cost as it does not involve explicit object detection or object-level feature extraction. However, global feature extraction with VideoMAE-V2 requires more FLOPs than other backbones. Nevertheless, the overall pipeline of VAGNet still requires fewer FLOPs per frame, even in the average case. The runtime analysis shows that despite the longer inference time of VideoMAE-V2, our overall VAGNet pipeline performs faster inference than others. VAGNet achieves 38 FPS on our limited computing platform, demonstrating potential for real-time inference on edge hardware. Another advantage of our method is that its computational cost is independent of the number of objects in the scene, which would enable it to maintain efficiency even in crowded urban environments. There are other methods that report higher FPS but on more powerful computing platforms. For example, GSC~\cite{wang2023gsc} and RARE~\cite{song2025real} report 58 FPS on an RTX 3090 GPU and 73 FPS on an RTX 6000 GPU, respectively. However, we note that VideoMAE-V2 has also achieved 112 FPS on a comparable high-end GPU, the RTX 4070~\cite{lee2026benchmarking}.



\begin{table}[htbp]
\scriptsize
\caption{FLOP counts of different methods. Counts for different stages are reported in GFLOPs/frame. The counts were obtained using the Python library THOP~\cite{THOP}.}
\centering
\setlength{\tabcolsep}{4pt}
\renewcommand{\arraystretch}{1.2}
\begin{tabular}{|c|c|c|c|c|}
\hline
\multirow{2}{*}{\textbf{Method}} & \multicolumn{4}{c|}{\textbf{GFLOPs }} \\ \cline{2-5}
 & \shortstack{\textbf{Object Detection}\\\textbf{\& Feature}\\ \textbf{Extraction}} 
 & \shortstack{\textbf{Global Feature}\\\textbf{Extraction}} 
 & \shortstack{\textbf{Model}\\\textbf{Inference}} 
 & \textbf{Total} \\
\hline

UString & \multirow{4}{*}{\shortstack{\\322.46 \\\\(worst case:\\19 objects/frame)}} & 15.47 (VGG16)    
& 10.49   & 348.42 \\ \cline{1-1}\cline{3-5}
Graph (Graph) & & 41.75 (I3D)           
& 2.14   & 366.35 \\ \cline{1-1}\cline{3-5}
Zhong~\cite{zhongearly2025} & & 41.75 (I3D)       
& 2.05   & 366.26 \\ \cline{1-1}\cline{3-5}
AAT-DA        &                        & 15.47 (VGG16)
& 1.05  & 461.96* \\ \cline{1-1}\cline{3-5}
STAGNet       &                        & 50.58 (SlowFast)     
& 2.20   & 375.24 \\ \cline{1-1}\cline{3-5}
\hline

UString & \multirow{4}{*}{\shortstack{\\123.25 \\\\(average case:\\6.12 objects/frame)}} & 15.47 (VGG16)    
& 10.49   & 149.21 \\ \cline{1-1}\cline{3-5}
Graph (Graph) & & 41.75 (I3D)           
& 2.14   & 167.14 \\ \cline{1-1}\cline{3-5}
Zhong~\cite{zhongearly2025} & & 41.75 (I3D)       
& 2.05   & 167.05 \\ \cline{1-1}\cline{3-5}
AAT-DA        &                        & 15.47 (VGG16)   
& 1.05  & 262.67* \\ \cline{1-1}\cline{3-5}
STAGNet       &                        & 50.58 (SlowFast)     
& 2.20   & 176.03 \\ \cline{1-1}\cline{3-5}
\hline
\multirow{4}{*}{VAGNet}     &       \multirow{4}{*}{\textbf{0}}         & 101.85 (VideoMAE)     
& \multirow{4}{*}{\textbf{0.033}}   & \textbf{101.88} \\
\cline{3-3}\cline{5-5}
      &                & 50.58 (SlowFast)     
&    & \textbf{50.51} \\
\cline{3-3}\cline{5-5}
      &                & 41.75 (I3D)     
&    & \textbf{41.78} \\
\cline{3-3}\cline{5-5}
      &                & 15.47 (VGG16)     
&    & \textbf{15.50} \\
\hline
\end{tabular}
\label{tab:execution_flops}
\begin{flushleft}
\scriptsize \textit{*} The total includes the 122.98 GFLOPs used for saliency/driver-attention computation. 
\end{flushleft}
\vspace{-4pt}
\end{table}

\begin{table}[htbp]
\scriptsize
\caption{The computational efficiency of the proposed and baseline methods is evaluated in terms of execution time. Module-level performance is reported in seconds per frame (s/f), whereas overall inference efficiency is reported as frames per second (FPS). For AAT-DA, the reported inference time additionally accounts for the generation of driver attention maps. Note that although we do not report the inference time for the method proposed by Zhong et. al~\cite{zhongearly2025} due to the unavailability of its implementation, we expect it to operate close to 23 FPS on our platform as it uses VGG16 and I3D as feature extractors.}
\centering
\setlength{\tabcolsep}{5pt}
\renewcommand{\arraystretch}{1.2}
\begin{tabular}{|c|c|c|c|c|}
\hline
\shortstack{\textbf{Method}} &
\shortstack{\textbf{Object}\\\textbf{ Detection}\\\textbf{\& Feature}\\\textbf{Extraction (s/f)}} &
\shortstack{\textbf{Global Feature}\\\textbf{Extraction (s/f)}} &
\shortstack{\textbf{Inference}\\\textbf{(s/f)}} &
\shortstack{\textbf{Overall}\\\textbf{speed}\\\\\textbf{(FPS)}} \\
\hline
UString       & \multirow{4}{*}{0.032}  & 0.004 (VGG16)     & 0.024        & 17 \\
Graph(Graph)  &                         & 0.01   (I3D)       & 0.001        & 23 \\
AAT-DA        &                         & 0.004 (VGG16)     & 0.094* & 8 \\
STAGNet       &                         & 0.014 (SlowFast)  & 0.001        & 21 \\
\hline
VAGNet    &    0                    & 0.025 (VideoMAE) & 0.001        & \textbf{38} \\
\hline
\end{tabular}
\label{tab:execution_fps}
\begin{flushleft}
\scriptsize \textit{*} 0.009 s for model inference + 0.085 s for generating the saliency (driver-attention) map.
\end{flushleft}
\vspace{-4pt}
\end{table}

\section{Limitations and future work}

Our model (and most models in the accident anticipation literature) uses dash-cam accident datasets curated from online sources and vehicle fleets. While these promote model generalization, as they include accident videos collected from many different ego-vehicles and dash-cams and cover various environmental and weather conditions, further improvement can be expected if the model can be finetuned after it has been deployed in a particular ego-vehicle. Such finetuning would allow the model to adapt to the dimensions and the viewpoint of the particular ego-vehicle. However, current models are not designed for post-deployment finetuning, as that would require accidents involving the vehicle they are deployed in\textemdash the very incidents that the models try to prevent. Therefore, future research can explore methods to finetune models based on normal driving footage (negative samples) captured during routine driving after model deployment.

Furthermore, although our approach requires fewer overall computations and less inference time per frame than existing methods, the VideoMAE-V2 backbone (base version) itself requires nearly 102 GFLOPs per frame. Future work can investigate knowledge distillation strategies for VideoMAE-V2 or other suitable backbones in the context of accident anticipation to further improve efficiency while preserving competitive accuracy. 

\section{Conclusions}

In this work, we presented VAGNet, a deep learning model for anticipating accidents from dash-cam video based on global features. The model is computationally efficient, as it 
does not perform explicit object-level feature extraction, but does so implicitly. Moreover, it anticipates accidents more accurately and earlier than existing methods, as demonstrated through experiments on four benchmark datasets.
Ablation studies reveal that the improved performance of the VAGNet model can be attributed to its transformer and graph-based architecture and the use of rich global features extracted through the VideoMAE-V2 foundation model.

\normalsize
\bibliography{Vision_CV}

@misc{shi2021maskedlabelpredictionunified,
      title={Masked Label Prediction: Unified Message Passing Model for Semi-Supervised Classification}, 
      author={Yunsheng Shi and Zhengjie Huang and Shikun Feng and Hui Zhong and Wenjin Wang and Yu Sun},
      year={2021},
      eprint={2009.03509},
      archivePrefix={arXiv},
      primaryClass={cs.LG},
      url={https://arxiv.org/abs/2009.03509}, 
}

@article{malawade2022spatiotemporal,
  title={Spatiotemporal scene-graph embedding for autonomous vehicle collision prediction},
  author={Malawade, Arnav Vaibhav and Yu, Shih-Yuan and Hsu, Brandon and Muthirayan, Deepan and Khargonekar, Pramod P and Al Faruque, Mohammad Abdullah},
  journal={IEEE Internet of Things Journal},
  volume={9},
  number={12},
  pages={9379--9388},
  year={2022},
  publisher={IEEE}
}

@article{yao2022dota,
  title={{DoTA: unsupervised detection of traffic anomaly in driving videos}},
  author={Yao, Yu and Wang, Xizi and Xu, Mingze and Pu, Zelin and Wang, Yuchen and Atkins, Ella and Crandall, David},
  journal={IEEE transactions on pattern analysis and machine intelligence},
  year={2022},
  publisher={IEEE}
}

@inproceedings{yu2020bdd100k,
  title={{BDD100K: A Diverse Driving Dataset for Heterogeneous Multitask Learning}},
  author={Yu, Fisher and Chen, Haofeng and Wang, Xin and Xian, Wenqi and Chen, Yingying and Liu, Fangchen and Madhavan, Vashisht and Darrell, Trevor},
  booktitle={Proceedings of the IEEE/CVF conference on computer vision and pattern recognition},
  pages={2636--2645},
  year={2020}
}

@article{yoffie2014mobileye,
  title={Mobileye: The future of driverless cars},
  author={Yoffie, David B},
  journal={Harvard Business School Case},
  pages={715--421},
  year={2014},
  publisher={Business History}
}

@article{ren2015faster,
  title={{Faster R-CNN: Towards Real-Time Object Detection with Region Proposal Networks}},
  author={Ren, Shaoqing and He, Kaiming and Girshick, Ross and Sun, Jian},
  journal={Advances in neural information processing systems},
  volume={28},
  year={2015}
}

@article{mahmood2023new,
  title={A New Approach to Traffic Accident Anticipation With Geometric Features for Better Generalizability},
  author={Mahmood, Farhan and Jeong, Daehyeon and Ryu, Jeha},
  journal={IEEE Access},
  volume={11},
  pages={29263--29274},
  year={2023},
  publisher={IEEE}
}

@article{Feichtenhofer2018SlowFastNF,
  title={{SlowFast Networks for Video Recognition}},
  author={Christoph Feichtenhofer and Haoqi Fan and Jitendra Malik and Kaiming He},
  journal={2019 IEEE/CVF International Conference on Computer Vision (ICCV)},
  year={2018},
  pages={6201-6210},
  url={https://api.semanticscholar.org/CorpusID:54463801}
}

@inproceedings{bao2020uncertainty,
  title={Uncertainty-based traffic accident anticipation with spatio-temporal relational learning},
  author={Bao, Wentao and Yu, Qi and Kong, Yu},
  booktitle={Proceedings of the 28th ACM International Conference on Multimedia},
  pages={2682--2690},
  year={2020}
}

@inproceedings{chan2017anticipating,
  title={Anticipating accidents in dashcam videos},
  author={Chan, Fu-Hsiang and Chen, Yu-Ting and Xiang, Yu and Sun, Min},
  booktitle={Computer Vision--ACCV 2016: 13th Asian Conference on Computer Vision, Taipei, Taiwan, November 20-24, 2016, Revised Selected Papers, Part IV 13},
  pages={136--153},
  year={2017},
  organization={Springer}
}

@article{simonyan2014very,
  title={Very deep convolutional networks for large-scale image recognition},
  author={Simonyan, Karen and Zisserman, Andrew},
  journal={arXiv preprint arXiv:1409.1556},
  year={2014}
}

@article{karim2022dynamic,
  title={A dynamic spatial-temporal attention network for early anticipation of traffic accidents},
  author={Karim, Muhammad Monjurul and Li, Yu and Qin, Ruwen and Yin, Zhaozheng},
  journal={IEEE Transactions on Intelligent Transportation Systems},
  volume={23},
  number={7},
  pages={9590--9600},
  year={2022},
  publisher={IEEE}
}

@article{karim2022toward,
  title={Toward explainable artificial intelligence for early anticipation of traffic accidents},
  author={Karim, Muhammad Monjurul and Li, Yu and Qin, Ruwen},
  journal={Transportation research record},
  volume={2676},
  number={6},
  pages={743--755},
  year={2022},
  publisher={SAGE Publications Sage CA: Los Angeles, CA}
}

@inproceedings{selvaraju2017grad,
  title={{Grad-CAM: Visual Explanations from Deep Networks via Gradient-based Localization}},
  author={Selvaraju, Ramprasaath R and Cogswell, Michael and Das, Abhishek and Vedantam, Ramakrishna and Parikh, Devi and Batra, Dhruv},
  booktitle={Proceedings of the IEEE international conference on computer vision},
  pages={618--626},
  year={2017}
}

@inproceedings{suzuki2018anticipating,
  title={Anticipating traffic accidents with adaptive loss and large-scale incident DB},
  author={Suzuki, Tomoyuki and Kataoka, Hirokatsu and Aoki, Yoshimitsu and Satoh, Yutaka},
  booktitle={Proceedings of the IEEE conference on computer vision and pattern recognition},
  pages={3521--3529},
  year={2018}
}

@inproceedings{bao2021drive,
  title={Drive: Deep reinforced accident anticipation with visual explanation},
  author={Bao, Wentao and Yu, Qi and Kong, Yu},
  booktitle={Proceedings of the IEEE/CVF International Conference on Computer Vision},
  pages={7619--7628},
  year={2021}
}

@article{fang2023vision,
  title={Vision-based traffic accident detection and anticipation: A survey},
  author={Fang, Jianwu and Qiao, Jiahuan and Xue, Jianru and Li, Zhengguo},
  journal={IEEE Transactions on Circuits and Systems for Video Technology},
  year={2023},
  publisher={IEEE}
}

@inproceedings{fang2019dada,
  title={{DADA-2000: Can Driving Accident be Predicted by Driver Attention? Analyzed by A Benchmark}},
  author={Fang, Jianwu and Yan, Dingxin and Qiao, Jiahuan and Xue, Jianru and Wang, He and Li, Sen},
  booktitle={2019 IEEE Intelligent Transportation Systems Conference (ITSC)},
  pages={4303--4309},
  year={2019},
  organization={IEEE}
}

@article{saito2015precision,
  title={The precision-recall plot is more informative than the ROC plot when evaluating binary classifiers on imbalanced datasets},
  author={Saito, Takaya and Rehmsmeier, Marc},
  journal={PloS one},
  volume={10},
  number={3},
  pages={e0118432},
  year={2015},
  publisher={Public Library of Science San Francisco, CA USA}
}

@inproceedings{thakur2024graph,
  title={Graph (graph): A nested graph-based framework for early accident anticipation},
  author={Thakur, Nupur and Gouripeddi, PrasanthSai and Li, Baoxin},
  booktitle={Proceedings of the IEEE/CVF Winter Conference on Applications of Computer Vision},
  pages={7533--7541},
  year={2024}
}

@inproceedings{carreira2017quo,
  title={Quo vadis, action recognition? a new model and the kinetics dataset},
  author={Carreira, Joao and Zisserman, Andrew},
  booktitle={proceedings of the IEEE Conference on Computer Vision and Pattern Recognition},
  pages={6299--6308},
  year={2017}
}

@article{chitraranjan2025vision,
  title={Vision-Based Collision Warning Systems with Deep Learning: A Systematic Review},
  author={Chitraranjan, Charith and Vipulananthan, Vipooshan and Sritharan, Thuvarakan},
  journal={Journal of Imaging},
  volume={11},
  number={2},
  pages={64},
  year={2025}
}

@inproceedings{kumamoto2025aat,
  title={{AAT-DA: Accident Anticipation Transformer with Driver Attention}},
  author={Kumamoto, Yuto and Ohtani, Kento and Suzuki, Daiki and Yamataka, Minori and Takeda, Kazuya},
  booktitle={Proceedings of the Winter Conference on Applications of Computer Vision},
  pages={1142--1151},
  year={2025}
}

@misc{IRB_Retrofit,
  title = {{Retrofit Collision Warning System Gives Older Vehicles A Safety Boost}},
  howpublished = "\url{https://trid.trb.org/View/1574810}",
  year = {2018}, 
  note = "[Online; accessed 27-October-2025]"
}

@article{vaswani2017attention,
  title={Attention is all you need},
  author={Vaswani, Ashish and Shazeer, Noam and Parmar, Niki and Uszkoreit, Jakob and Jones, Llion and Gomez, Aidan N and Kaiser, {\L}ukasz and Polosukhin, Illia},
  journal={Advances in neural information processing systems},
  volume={30},
  year={2017}
}

@ARTICLE{vipulananthan2025stagnet,
  author={Vipulananthan, Vipooshan and Mohottala, Kumudu and Chinthana, Kavindu and Paramulla, Nimsara and Chitraranjan, Charith},
  journal={IEEE Access}, 
  title={{STAGNet: A Spatio-Temporal Graph and LSTM Framework for Accident Anticipation}}, 
  year={2025},
  volume={13},
  number={},
  pages={213769-213779},
  doi={10.1109/ACCESS.2025.3645127}}

@inproceedings{zhao2023gated,
  title={Gated driver attention predictor},
  author={Zhao, Tianci and Bai, Xue and Fang, Jianwu and Xue, Jianru},
  booktitle={2023 IEEE 26th International Conference on Intelligent Transportation Systems (ITSC)},
  pages={270--276},
  year={2023},
  organization={IEEE}
}

@article{zhongearly2025,
  title={Early Traffic Accident Anticipation via Feature Consistency Representation and Soft Label Regression},
  author={Zhong, Yuanhong and Yan, Ge and Zhu, Ruyue and Gan, Ping and Shen, Xuerui},
  journal={ACM Transactions on Multimedia Computing, Communications and Applications},
  year={2025},
  publisher={ACM New York, NY}
}

@article{liu2025ccaf,
  title={{CCAF-Net: Cascade Complementarity-Aware Fusion Network for traffic accident prediction in dashcam videos}},
  author={Liu, Wei and Li, Yafei and Zhang, Tao and Gao, Yixiang and Wei, Longsheng and Chen, Jun},
  journal={Neurocomputing},
  volume={624},
  pages={129285},
  year={2025},
  publisher={Elsevier}
}

@article{fang2022cognitive,
  title={Cognitive accident prediction in driving scenes: A multimodality benchmark},
  author={Fang, Jianwu and Li, Lei-Lei and Yang, Kuan and Zheng, Zhedong and Xue, Jianru and Chua, Tat-Seng},
  journal={arXiv preprint arXiv:2212.09381},
  year={2022}
}

@article{li2024cognitive,
  title={Cognitive traffic accident anticipation},
  author={Li, Lei-Lei and Fang, Jianwu and Xue, Jianru},
  journal={IEEE Intelligent Transportation Systems Magazine},
  volume={16},
  number={5},
  pages={17--32},
  year={2024},
  publisher={IEEE}
}

@inproceedings{moura2025nexar,
  title={Nexar Dashcam Collision Prediction Dataset and Challenge},
  author={Moura, Daniel and Zhu, Shizhan and Zvitia, Orly},
  booktitle={Proceedings of the Computer Vision and Pattern Recognition Conference},
  pages={2583--2591},
  year={2025}
}

@inproceedings{wang2023videomae,
  title={Videomae v2: Scaling video masked autoencoders with dual masking},
  author={Wang, Limin and Huang, Bingkun and Zhao, Zhiyu and Tong, Zhan and He, Yinan and Wang, Yi and Wang, Yali and Qiao, Yu},
  booktitle={Proceedings of the IEEE/CVF conference on computer vision and pattern recognition},
  pages={14549--14560},
  year={2023}
}

@article{tekkesinoglu2025advancing,
  title={Advancing explainable autonomous vehicle systems: A comprehensive review and research roadmap},
  author={Tekkesinoglu, Sule and Habibovic, Azra and Kunze, Lars},
  journal={ACM Transactions on Human-Robot Interaction},
  volume={14},
  number={3},
  pages={1--46},
  year={2025},
  publisher={ACM New York, NY}
}

@article{yu2021scene,
  title={Scene-graph augmented data-driven risk assessment of autonomous vehicle decisions},
  author={Yu, Shih-Yuan and Malawade, Arnav Vaibhav and Muthirayan, Deepan and Khargonekar, Pramod P and Al Faruque, Mohammad Abdullah},
  journal={IEEE Transactions on Intelligent Transportation Systems},
  volume={23},
  number={7},
  pages={7941--7951},
  year={2021},
  publisher={IEEE}
}

@article{liu2025seeing,
  title={Seeing before Observable: Potential Risk Reasoning in Autonomous Driving via Vision Language Models},
  author={Liu, Jiaxin and Yan, Xiangyu and Peng, Liang and Yang, Lei and Zhang, Lingjun and Luo, Yuechen and Tao, Yueming and Tan, Ashton Yu Xuan and Li, Mu and Zhang, Lei and others},
  journal={arXiv preprint arXiv:2511.22928},
  year={2025}
}

@article{mattas2022driver,
  title={Driver models for the definition of safety requirements of automated vehicles in international regulations. Application to motorway driving conditions},
  author={Mattas, Konstantinos and Albano, Giovanni and Don{\`a}, Riccardo and Galassi, Maria Christina and Suarez-Bertoa, Ricardo and Vass, Sandor and Ciuffo, Biagio},
  journal={Accident Analysis \& Prevention},
  volume={174},
  pages={106743},
  year={2022},
  publisher={Elsevier}
}

@article{zhang2026intelligent,
  title={Intelligent defensive driving for autonomous vehicles: Framework, strategy and verification},
  author={Zhang, Ting and Wang, Zixuan and Wang, Hong and Li, Jun},
  journal={Accident Analysis \& Prevention},
  volume={226},
  pages={108355},
  year={2026},
  publisher={Elsevier}
}

@article{li2024smpc,
  title={{SMPC-Based Motion Planning of Automated Vehicle When Interacting With Occluded Pedestrians}},
  author={Li, Daofei and Jiang, Yangye and Zhang, Jiajie and Xiao, Bin},
  journal={IEEE Transactions on Intelligent Transportation Systems},
  year={2024},
  publisher={IEEE}
}

@article{mannering2016unobserved,
  title={Unobserved heterogeneity and the statistical analysis of highway accident data},
  author={Mannering, Fred L and Shankar, Venky and Bhat, Chandra R},
  journal={Analytic methods in accident research},
  volume={11},
  pages={1--16},
  year={2016},
  publisher={Elsevier}
}

@misc{WHO_accidents,
  title = {{Road traffic injuries}},
  howpublished = "\url{https://www.who.int/news-room/fact-sheets/detail/road-traffic-injuries}",
  year = {2023}, 
  note = "[Online; accessed 24-December-2025]"
}

@article{liu2024curse,
  title={Curse of rarity for autonomous vehicles},
  author={Liu, Henry X and Feng, Shuo},
  journal={nature communications},
  volume={15},
  number={1},
  pages={4808},
  year={2024},
  publisher={Nature Publishing Group UK London}
}

@article{chib2023recent,
  title={Recent advancements in end-to-end autonomous driving using deep learning: A survey},
  author={Chib, Pranav Singh and Singh, Pravendra},
  journal={IEEE Transactions on Intelligent Vehicles},
  volume={9},
  number={1},
  pages={103--118},
  year={2023},
  publisher={IEEE}
}

@article{zhang2023deep,
  title={Deep long-tailed learning: A survey},
  author={Zhang, Yifan and Kang, Bingyi and Hooi, Bryan and Yan, Shuicheng and Feng, Jiashi},
  journal={IEEE transactions on pattern analysis and machine intelligence},
  volume={45},
  number={9},
  pages={10795--10816},
  year={2023},
  publisher={IEEE}
}

@article{song2024dynamic,
  title={Dynamic attention augmented graph network for video accident anticipation},
  author={Song, Wenfeng and Li, Shuai and Chang, Tao and Xie, Ke and Hao, Aimin and Qin, Hong},
  journal={Pattern Recognition},
  volume={147},
  pages={110071},
  year={2024},
  publisher={Elsevier}
}

@misc{THOP,
  title = {{THOP: A tool to count the FLOPs of PyTorch model.}},
  howpublished = "\url{https://pypi.org/project/thop/}",
  year = {2022}, 
  note = "[Online; accessed 14-November-2025]"
}

@article{wang2023gsc,
  title={{GSC: A Graph and Spatio-Temporal Continuity Based Framework for Accident Anticipation}},
  author={Wang, Tianhang and Chen, Kai and Chen, Guang and Li, Bin and Li, Zhijun and Liu, Zhengfa and Jiang, Changjun},
  journal={IEEE Transactions on Intelligent Vehicles},
  volume={9},
  number={1},
  pages={2249--2261},
  year={2023},
  publisher={IEEE}
}

@article{simeoni2025dinov3,
  title={Dinov3},
  author={Sim{\'e}oni, Oriane and Vo, Huy V and Seitzer, Maximilian and Baldassarre, Federico and Oquab, Maxime and Jose, Cijo and Khalidov, Vasil and Szafraniec, Marc and Yi, Seungeun and Ramamonjisoa, Micha{\"e}l and others},
  journal={arXiv preprint arXiv:2508.10104},
  year={2025}
}

@inproceedings{song2025real,
  title={Real-time traffic accident anticipation with feature reuse},
  author={Song, Inpyo and Lee, Jangwon},
  booktitle={2025 IEEE International Conference on Image Processing (ICIP)},
  pages={2312--2317},
  year={2025},
  organization={IEEE}
}

@article{wang2024yolov10,
  title={Yolov10: Real-time end-to-end object detection},
  author={Wang, Ao and Chen, Hui and Liu, Lihao and Chen, Kai and Lin, Zijia and Han, Jungong and Ding, Guiguang},
  journal={Advances in neural information processing systems},
  volume={37},
  pages={107984--108011},
  year={2024}
}

@inproceedings{he2016deep,
  title={Deep residual learning for image recognition},
  author={He, Kaiming and Zhang, Xiangyu and Ren, Shaoqing and Sun, Jian},
  booktitle={Proceedings of the IEEE conference on computer vision and pattern recognition},
  pages={770--778},
  year={2016}
}

@article{tatar2023real,
  title={Real-time multi-task ADAS implementation on reconfigurable heterogeneous MPSoC architecture},
  author={Tatar, Guner and Bayar, Salih},
  journal={IEEE Access},
  volume={11},
  pages={80741--80760},
  year={2023},
  publisher={IEEE}
}

@article{nguyen2024optimizing,
  title={Optimizing monocular driving assistance for real-time processing on Jetson AGX Xavier},
  author={Nguyen, Huy-Hung and Tran, Duong Nguyen-Ngoc and Pham, Long Hoang and Jeon, Jae Wook},
  journal={IEEE Access},
  volume={12},
  pages={71853--71865},
  year={2024},
  publisher={IEEE}
}

@article{falaschetti2024embedded,
  title={Embedded real-time vehicle and pedestrian detection using a compressed tiny YOLO v3 architecture},
  author={Falaschetti, Laura and Manoni, Lorenzo and Palma, Lorenzo and Pierleoni, Paola and Turchetti, Claudio},
  journal={IEEE Transactions on Intelligent Transportation Systems},
  volume={25},
  number={12},
  pages={19399--19414},
  year={2024},
  publisher={IEEE}
}

@article{guan2022lightweight,
  title={A lightweight framework for obstacle detection in the railway image based on fast region proposal and improved YOLO-tiny network},
  author={Guan, Ling and Jia, Limin and Xie, Zhengyu and Yin, Chaoying},
  journal={IEEE Transactions on Instrumentation and Measurement},
  volume={71},
  pages={1--16},
  year={2022},
  publisher={IEEE}
}

@article{lee2026benchmarking,
  title={Benchmarking action recognition models for self-harm detection in studio and real-world datasets},
  author={Lee, Kanghee and Lee, Doohee and Ham, Hyun-Sik and Kim, Hee-Cheol and Lee, Yourack and Jeong, Hyun-Ghang and Choi, Hyun-Soo},
  journal={Scientific Reports},
  year={2026},
  publisher={Nature Publishing Group UK London}
}

\end{document}